\documentclass[10pt, a4paper]{article}
\usepackage{lrec2016}
\usepackage{multibib}
\newcites{languageresource}{Language Resources}
\usepackage{graphicx}

\usepackage{epstopdf}

\usepackage[utf8]{inputenc}  
\usepackage{times}
\usepackage{url}
\usepackage{latexsym}
\usepackage{multirow}
\usepackage{color}
\usepackage{pbox}
\usepackage{lipsum}
\usepackage{booktabs}

\title{SweLL on the rise: \protect\\ Swedish Learner Language corpus for European Reference Level studies}

\name{Elena Volodina$^1$, Ildikó Pil\'{a}n$^1$, Ingegerd Enström$^1$, Lorena Llozhi$^1$, \protect\\
\textbf{\large Peter Lundkvist$^2$, Gunlög Sundberg$^2$, Monica Sandell$^3$}}

\address{$^1$ Spr\aa{}kbanken, University of Gothenburg, Sweden \\
         $^2$ Department of Swedish Language and Multilingualism, Stockholm University, Sweden \\
         $^3$ Center for Language Introduction, Gothenburg, Sweden \\
         elena.volodina@svenska.gu.se \\}

\abstract{We present a new resource for Swedish, SweLL, a corpus of Swedish Learner essays linked to learners’ performance according to the Common European Framework of Reference (CEFR). SweLL consists of three subcorpora – SpIn, SW1203 and Tisus, collected from three different educational establishments. The common metadata for all subcorpora includes age, gender, native languages, time of residence in Sweden, type of written task. Depending on the subcorpus, learner texts may contain additional information, such as text genres, topics, grades.  Five of the six CEFR levels are represented in the corpus: A1, A2, B1, B2 and C1 comprising in total 339 essays. C2 level is not included since courses at C2 level are not offered. The work flow consists of collection of essays and permits, essay digitization and registration, meta-data annotation, automatic linguistic annotation. Inter-rater agreement is presented on the basis of SW1203 subcorpus. The work on SweLL is still ongoing with more that 100 essays waiting in the pipeline. This article both describes the resource and the “how-to” behind the compilation of SweLL.
\newline
\newline  }

\begin{document}

\maketitleabstract

\section{Introduction}
\label{introduction}

With globalization and a growing number of people seeking asylum, better work or better living conditions in Europe in general and in Sweden in our particular case \cite{migrationsverket2014}, the need for effective foreign and second language (L2) teaching and analysis of L2 is in every way important. Access to digitized samples of language that L2 learners produce can facilitate L2 research into usage of various linguistic aspects, constructions and competences, move forward development of methods for automatic L2 analysis and become a way to optimize both teaching and creation of new learning tools and materials.

The SweLL corpus presented here is a collection of learner-written essays at different proficiency levels, as defined by the Common European Framework of Reference \cite{councilofeurope2001}. As such, the corpus is an evidence of learner language that facilitates research on interlanguage \cite{selinker1972}, which is a term for a dynamically developing system that L2 learners build before they become proficient in the target language. Interlanguage has been a focus of much linguistic and pedagogical research over the past decades \cite{selinker1972}, whereas computational linguistic methods for interlanguage analysis only recently have started to gain international attention and are currently explored for languages that offer electronic access to annotated learner data, such as essays and speech transcripts \cite{rosenetal2014}.

Research on L2 acquisition and learning was for long based on assumptions of language, rather than on the study of learners' developing language, or interlanguage. Empirical studies on interlanguage have been carried out since the late 1960s, but much research has been based on smaller scale studies of specific structures. This is especially true for Swedish; while L2 corpora have been available for e.g. English and Norwegian for the past 2-3 decades, resources for this kind of studies have been largely lacking for L2 Swedish. However, researchers of L2 vocabulary and grammar acquisition are in great demand of digitized L2 corpora of Swedish, that can help verify hypotheses generated by experimental studies and/or smaller scaled empirical studies. This is also true for those who pursue research on structures in-between grammar and lexicon, captured by construction grammar \cite{skoldbergetal2013}, which is, by definition, usage-based and internationally increasingly concerned with L2 learner perspectives.

\begin{figure}
\fbox{\parbox{8cm}{Has \textit{sufficient vocabulary} to conduct routine, everyday transactions involving \textit{familiar situations and topics}. 

Has a sufficient vocabulary for the expression of \textit{basic communicative needs}. Has a sufficient vocabulary for coping with \textit{simple survival needs}.}}

\caption{CEFR descriptor for vocabulary range at A2 level \cite[112]{councilofeurope2001}. Subject to interpretations is \textit{sufficient vocabulary, familiar situations and topics, basic communicative needs, simple survival needs.} }
\label{cefrFig}
\end{figure}

\begin{table*}[t]
\begin{small}
\begin{center}

\begin{tabular}{|c|c|c|c|c|c|}
\hline \bf Subcorpus & \bf Levels  & \bf Period & \bf School & \bf Type of essays & \bf Size \\
\hline
SpIn 
& \pbox{3cm}{A1-B1} 
& \pbox{5cm}{2013-2015} 
& \pbox{7cm}{Center for Language \\ Introduction} 
& \pbox{15cm}{Mid-term exams, multiple topics} 
& \pbox{10cm}{144 essays / 85 students} \\[10pt] \hline
SW1203 
& \pbox{3cm}{B1-C1} 
& \pbox{5cm}{2012-2013} 
& \pbox{7cm}{University of Gothenburg} 
& \pbox{15cm}{3 essays written by each student for \\ entrance, intermediate and final exams } 
& \pbox{10cm}{90 essays / 52 students} \\[10pt] \hline
Tisus
& \pbox{3cm}{B2-C1} 
& \pbox{5cm}{2006-2007} 
& \pbox{7cm}{Stockholm University} 
& \pbox{15cm}{External (high-stake) exam, same topic} 
& \pbox{10cm}{105 essays / 105 students} \\[7pt] 
\hline
\end{tabular}
\medskip
\caption{\label{subcorpora} Overview over SweLL subcorpora}
\end{center}
\end{small}
\end{table*}

Moreover, this type of data is needed for the interpretation of the CEFR descriptors at each level of proficiency. The CEFR \cite{councilofeurope2001}, adopted in 2001, has been in need of further guidelines and specifications for each individual language since it is too vague by nature to cover a range of different languages \cite{byrnes2007,little2011common}. Figure 1 demonstrates an example of a CEFR descriptor where expressions in italics are very difficult to interpret in terms of vocabulary that a learner needs to acquire. Many of the languages that work with the CEFR specifications operate on the basis of digitized L2 learner production, e.g. Norwegian \cite{carlsen2010}, English \cite{hawkins2010}, Italian, German and Czech \cite{abeletal2014}. The efforts to interpret CEFR descriptors and “can-do” statements are being undertaken for a number of languages\footnote{see http://www.coe.int/t/dg4/linguistic/dnr\_EN.asp? for the \\ list of concerned languages} \cite{marello2012}, however, Swedish has not yet received much attention.

Over time, Swedish L2 learner essays have been collected in a number of projects and resulted in several learner corpora, e.g. ASU \cite{hammarberg2005}, CrossCheck \cite{lindberg2004}, Swedish EALA \cite{saxena2002}. None of the corpora are labeled with the CEFR proficiency levels, mostly since they have been collected before the Framework was accepted. Besides, these essay collections are available only for closed research groups and are not widely accessible for outside users for online use, which leaves a gap to be filled.


\section{SweLL data sources}
\label{sources}

SweLL stands for \textbf{Swe}dish \textbf{L}earner \textbf{L}anguage. It is a constantly growing corpus with a specific focus on CEFR-annotated essays. At the moment it contains three subcorpora covering five out of the six CEFR levels, excluding C2, as shown in Table 1. \textit{SpIn} part is constantly growing, with more than 100 essays in a pipeline for addition; collaboration with the Center on collection of future essays is ongoing. \textit{SW1203} contains in total 144 essays, where 35 students have written all the 3 essays. The missing 54 essays will be added to the collection shortly. The present \textit{Tisus} collection contains essays from Spring 2006, however, essays are written twice a year, and a large number of essays are  waiting to be added to the corpus.

\subsection{SpIn subcorpus}
\label{spin}

Over the period of 2013-2015, L2 essays have been collected from the Center for Language Introduction (Centrum för \textbf{Sp}råk\textbf{In}troduktion). The Center receives young L2 learners (16-20 years) - refugees and other immigrant groups - for a one-year intensive program that aims to prepare learners for the next transitional training stage before they can proceed with the upper-secondary studies at national Swedish schools. Most of students are absolute beginners, and depending on their study tempo and background education path, they can cover different number of levels during their time at the Center. Every 7 weeks students' abilities are tested with the aim to regroup students according to the achieved level of proficiency and demonstrated learning style and tempo. Among others, students write essays as a part of these tests. A selection of these essays are added to SweLL. Most of the essays have public permits\footnote{example of a permit: http://spraakbanken.gu.se/sites/spraak banken.gu.se/files/tillstand\_eng-24042013\_v03.pdf}, i.e. permits which allow the essays to be published for anyone to explore provided the essays are anonymized and do not reveal student identity.

A remarkable feature of SpIn is that there are a number of “returning” students (85 students and 144 essays) with several essays written over time, which makes it a perfect basis for tracking gradual development of various linguistic constructs. Another feature is that each essay is labeled with one or more topics according to the same taxonomy that has been used in COCTAILL \cite{volodinaetal2014}, a corpus of L2 coursebooks, which makes it possible to compare topic vocabulary in reading material (receptive language) versus written essays (productive language).

\subsection{SW1203 subcorpus}
\label{sw1203}

SW1203 is an acronym for a course \textit{Swedish as a foreign language - Qualifying course in Swedish}, a course given at the University of Gothenburg, that prepares foreign students for university studies in Sweden. It covers one term of full-time studies. To qualify for the course, an entrance exam has to be taken where an overall level of B2 should be demonstrated based on essay writing, grammar and vocabulary tests, and reading comprehension tasks. During the period of one term, students write a number of essays, of which three have been collected for the SW1203 corpus: the ones written for the entrance exam, for mid-term assessment, and for a final test. The same students thus write at least three essays, which provides a unique opportunity to trace linguistic development of students over time. SW1203 essays have been collected during 2012-2013, and have a restricted research permits. Additional information on topics and genres is available.

\subsection{Tisus subcorpus}
\label{tisus}

\begin{figure*}[t]
   \begin{center}
     \includegraphics[width=18cm]{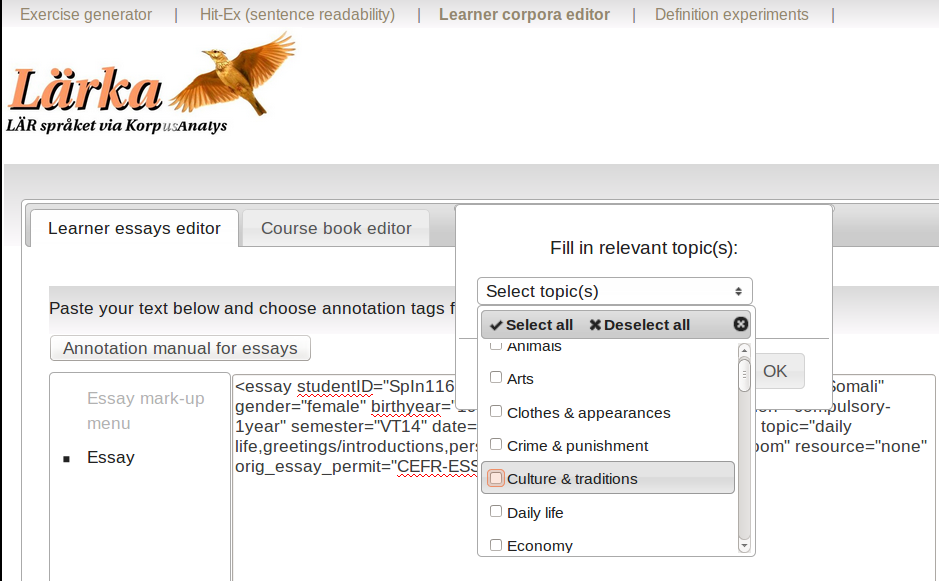}
     \caption{Learner essays editor for storing student profiles and generation of metadata annotation} 
     \label{editor} 
   \end{center} 
\end{figure*}

sTisus stands for \textbf{T}est \textbf{I}n \textbf{S}wedish for \textbf{U}niversity \textbf{S}tudies, a high-stakes exam taken by prospective university students to qualify for university studies. This exam consists of several parts, including written essays, an oral exam and a reading comprehension test. Holistic assessment, as well as analytic assessment per test construct are available for each essay. As all SweLL essays, Tisus-essays were hand-written and later digitized. Each essay has a restricted research permit\footnote{Restricted permit means that essays can be viewed in the browsing tool Korp by an approved group of researchers with password protection, and are not accessible to the general public.}. All essays are on the same topic (“Stress”) and within the same genre of argumentative writing, which figures in the essay metadata.

\section{SweLL workflow}
\label{workflow}

Essays have been collected from three academic institutions (Table \ref{subcorpora}) where both L2 teaching and L2 assessment are practised. Permits have been collected from students for each essay allowing either open or restricted research usage. SweLL collection facilitates browsing several essays written by the same student and thus follow progression in student’s language development over time, a feature that is seldom available in other L2 corpora collections.

\textit{CEFR-rating} of the essays was performed by a set of trained assessors, who are qualified language teachers, taking part in assessment training sessions each year. The normal practice has been to use several assessors, a minimum of two, for each essay, which gave us a possibility to calculate inter-annotator agreement, reported below for the SW1203 subcorpus. The assessment was performed with slightly different scales depending on the academic institution. Some of scales included, besides the level itself, grades of performance (e.g. for Tisus, grades of performance are assigned on the scale of 1-5 where 3-5 are pass grades, 5 being the highest one); others provided subtle subcategories (degrees) of the reached level (e.g. B1-, B1.1, B1.2, B1+). 

\textit{Inter-annotator agreement} is a degree to which several annotators agree about assigning certain attributes. It can be reported in a number of ways, depending upon the number of annotators (pairwise, multiple) and the number and type of values to be assigned. The pairwise agreement in terms of Krippendorff's alpha \cite{krippendorff1980} for assigning one of the five CEFR levels was 0.80 using ordinal distance, i.e. taking into consideration the \textit{degree of difference} between the annotated levels instead of a simple binary distinction of matching and non-matching assigned levels. This value reaches the threshold value specified in \cite{artstein2008} for assuring a good annotation quality.

Three major \textit{principles for essay digitization} have been applied: 
\begin{itemize}
\item not to reveal the author’s identity, where, for example, all revealing names and addresses have been substituted with \textit{NN} or \textit{NN-street}.
\item not to correct author’s mistakes. However, in dubious cases, we applied a principle of \textit{positive assumption}, i.e. that learners meant the correct variant. 
\item not to hypothesize when handwriting is illegible, where each illegible letter was substituted with @, and stricken-out text was left out.
\end{itemize}

As for \textit{meta-information}, for each of the subcorpora, we have collected 
\begin{itemize}
\item \textit{learner variables}: age at the moment of writing, gender, mother tongue (L1), education level, residence time in Sweden;
\item \textit{essay-related information}: assigned CEFR level, setting (exam/classroom/home), access to extra materials (e.g. lexicons, statistics), academic term and date when the essays have been written, essay title, and depending upon the subcorpus - topics (SpIn, TISUS, SW1203), genre (TISUS, SW1203), grade (TISUS)
\end{itemize}

To facilitate consistency in \textit{metadata markup}, a special editor\footnote{http://spraakbanken.gu.se/larka/larka\_cefr\_editor.html} has been implemented (Figure \ref{editor}) on the basis of Lärka, a language learning platform \cite{volodina2014}, where essay-related information is automatically stored to a server. An annotator is steered through prompts to fill in or values to select from. A studentID and an essayID are automatically suggested, whereas if a student profile already exists on the server, the essayID will reflect that. Once all information is provided, a resulting xml-tag is generated.

\textit{Linguistic annotation} has been automatically added to the corpus using Korp pipeline \cite{borinetal2012}, including lemmatization, Part-Of-Speech-tagging and syntactic information. While we are aware of the infelicities of the learner language and the effect it can have on the quality of annotation, that has been so far the only way of dealing with learner essays. In the future, SweLL corpus will be used to normalize learner essays and create methods for dealing with deviations of L2 learner language.

All subcorpora are available for browsing through Korp\footnote{Korp is a corpus infrastructure of the Swedish Language Bank: www.spraakbanken.gu.se/korp}, with or without password protection depending upon the permit type. Special support for specific browsing of learner essays (such as browsing full essays as opposed to KWIC-mode) is, however, not yet available pending sufficient funding.


\section{SweLL in numbers}
\label{statistics}

The corpus consists of 339 essays, comprising in total 9~373 sentences and 144~087 tokens. The size of SweLL per subcorpus is presented in Table \ref{swell_size}.

\begin{table}[h]
\begin{center}
\begin{tabular}{|c|c|c|c|}
\hline \bf Subcorpus & \bf Nr essays & \bf Nr sentences & \bf Nr tokens\\
\hline
\bf Tisus & 105 & 3 367 & 59 213 \\
\bf Sw1203 & 90 & 3 153 & 52 017 \\
\bf SpIn & 144 & 2 853 & 32 857 \\
\bf Total & 339 & 9 373 & 144 087\\

\hline
\end{tabular}
\medskip
\caption{\label{swell_size} The size of SweLL}
\end{center}
\end{table}

The distribution of the essays across levels is somewhat unbalanced for the moment\footnote{About 30 additional essays have recently become available for A1 level which will be soon included in our corpus}, the number of A1-level essays is almost a fifth less than those of all other CEFR levels as Table \ref{size_level} shows. This may be because at the initial stage of language learning, students’ knowledge of L2 Swedish is not yet mature enough to write essays often and other, less complex tasks are preferred. Besides, the number of classroom hours to complete A1 level is usually much shorter than for higher levels of proficiency.

\begin{table}[h]
\begin{center}
\begin{tabular}{|c|c|c|c|c|c|c|c|}
\hline \bf Sub- &&&&&& \bf Un- & \\
\bf corpus & \bf A1 & \bf A2 & \bf B1 & \bf B2 & \bf C1 &\bf known& \bf Total\\
\hline
\bf Tisus & - & - & - & 27 & 78 & - & 105 \\
\hline
\bf Sw1203 &-& - & 33 & 45 & 11 & 1 & 90 \\
\hline
\bf SpIn & 16  & 83 & 42 & 2 & -& 1& 144 \\
\hline
\bf Total & 16 & 83 & 75 & 74 & 89 & 2 & \bf 339 \\
\hline
\end{tabular}
\medskip
\caption{\label{size_level} Number of essays per CEFR level and subcorpus}
\end{center}
\end{table}

In the case of 45 students, more than one essay is available which could provide interesting material for learner profiling. The SweLL collection contains 12 unique students who have authored 4 or 5 essays, 22 students with 3 essays, 11 students with 2 essays, and the remaining 156 unique students with 1 essay. Since essays contain information about the date of the essay, it is possible to follow learner's linguistic development over time in the case of the 45 multiple-essay authors. 

\begin{figure}[h]
   \begin{center}
     \includegraphics[width=8.5cm]{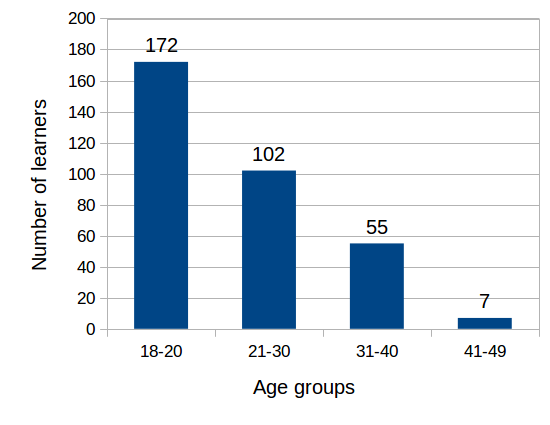}
     \caption{Learners by age groups} 
     \label{age} 
   \end{center} 
\end{figure}

As for learners’ gender, female writers are somewhat more represented than male ones, authoring 60 \% of the essays. Learners’ age spans from 16 to 49 years, with the predominant age span between 18 and 20 years. From the statistics in Figure \ref{age} it becomes obvious that the older age groups do not as eagerly engage in learning Swedish as their second language compared to younger people.

\begin{table*}[t]
\begin{center}
\begin{tabular}{|c|c|p{4.2cm}|p{4.2cm}|p{4.2cm}|}
\hline
\bf CEFR & \bf Source & \bf Original sentence & \bf Normalized & \bf Translation \\
\hline
A1 & SpIn &\textit{eter lunch går hem.}& Jag äter lunch och går hem.
&`I eat lunch and go home.'\\
\hline
A2 & SpIn & \textit{min pappa och jag äter frukost i köket.} & Min pappa och jag äter frukost i köket. &`My dad and I eat breakfast in the kitchen.'\\
\hline
B1 &SpIn&\textit{Läkare sa att jag måste äta bra om jag ville blir frisk.} & Läkaren sa att jag måste äta väl om jag ville bli frisk.&`The doctor said that I have to eat well if I wanted to get better.'\\
\hline
B2 & Sw1203 &\textit{Någon kan sitta på en bra restaurang, ätta dyr mat och fortfarande känna sig dålig.} & Någon kan sitta på en bra restaurang, äta dyr mat och fortfarande känna sig dåligt. &`One can sit in a good restaurant, eat expensive food and still feel sick.'\\
\hline
C1 &Tisus&\textit{I dagens samhälle är det inte längre en självklarhet att man har tid att äta .} & &`In today's society, it is not obvious any more that one has time to eat.'\\
\hline
\end{tabular}
\caption{\label{table:eg_sents} Example sentences per CEFR level.}
\end{center}
\end{table*}

Due to the essay topic markup in two of the three subcorpora, it is possible to study topic-related linguistic variables, such as vocabulary. The following topics are present in SpIn and Tisus, listed below from most frequent to the least frequent ones:

\begin{itemize}
    \item health and body care: 	117 essays
	\item personal identification: 	97
	\item daily life: 	60
	\item relations with other people: 	31
	\item free time, entertainment: 	19
	\item places: 	16
	\item arts: 	15
	\item travel: 	15
	\item education: 	9
	\item family and relatives: 	7
	\item economy 	4
\end{itemize}

\begin{figure}[h]
   \begin{center}
     \includegraphics[width=9.5cm]{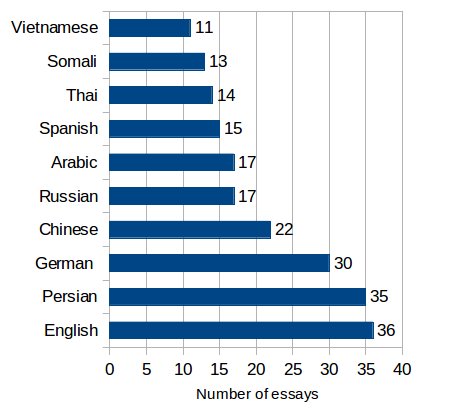}
     \caption{Distribution of learners’ native languages} 
     \label{l1} 
   \end{center} 
\end{figure}

Learners in SweLL have a wide variety of native languages, amounting to a total of 64 languages from different language families. The distribution of the ten most frequent native languages is presented in Figure \ref{l1}.

\begin{figure}[h]
   \begin{center}
     \includegraphics[width=8.5cm]{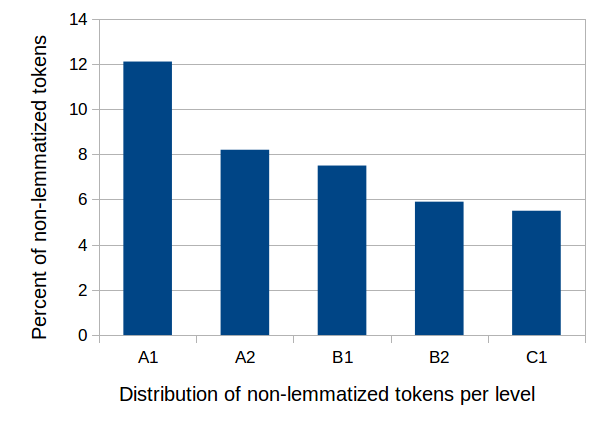}
     \caption{Distribution of non-lemmatized tokens, given in \% per level} 
     \label{non-words} 
   \end{center} 
\end{figure}

An interesting piece of statistics is the number of non-lemmatized words per level. Hypothetically, most of the running text words that couldn't be automatically matched with entries in a lexicon during lemmatization are either misspelled or non-existent words. The number thus might be assumed to show the lexical/orthographical error rate. At A1 level the percentage of such items is 4\% higher than at A2, and decreases by further 2.5\% at C1 level, as shown in Figure \ref{non-words}.

Finally, in Table \ref{table:eg_sents} we present an example sentence containing the verb \textit{äta} `eat' from an essay for each CEFR level. Each sentence is followed by a normalized (error-corrected) version in case the original sentence was incorrect, as well as an English translation.

\section{Concluding remarks and future prospects}

In SweLL, we have collected essays written by L2 learners of Swedish, in order to use them, on the one hand, to develop (semi-)automatic methods for L2 analysis and annotation, and, on the other, to offer access to L2 empiric data to all interested target groups, such as linguists, L2 researchers, teachers and students. 

The SweLL corpus is not yet finalized. The essay collection is an ongoing process and we plan to periodically extend SweLL with new essays.

Depending upon funding, several further steps are envisaged for SweLL development: 
\begin{enumerate}
\item to add normalized versions along the learner-written versions (in a parallel corpus fashion); 
\item to add error annotation; 
\item on a broader front, to develop methods for the automatic analysis and annotation of L2 written production. 
\end{enumerate}

While (1) and (2) will facilitate the creation of a \textit{gold standard corpus} for L2 Swedish, (3) will exploit the result of that. Relevance of (3) comes with the fact that existing computational linguistic methods for text annotation are developed with a normative language in mind, e.g. well-written newspaper texts, and cannot be applied effectively in their current form to L2 texts. However, annotating learner data manually is an extremely time-consuming and costly enterprise. To cater for the grammatical and orthographical infelicities in L2 texts, and to make annotation of L2 data more time-effective, computational linguistic methods need to be \textbf{adapted to the challenges set by interlanguage}. By developing these methods, we can bring forward research within computational linguistics since it lacks “methods targeting learner texts” \cite{rosenetal2014}, as well as can facilitate growth of Swedish L2 data thus paving the way for corpus-based L2 research on Swedish.

Multiple linguistic and pedagogical exploitation scenarios can be envisaged given that (1) and (2) above are available, such as to search for all (mis)spelling variants of some lemma, e.g. “mycket” (“much”) and get hits with all variations “mycekt*”, “miket*”, “micke*”. Another example is to trace (in)correct use of possessive constructions in essays written by the same student over time, or students sharing the same mother tongue, and get results showing types and percentage of erroneous/correct use at the beginner level (e.g. \textit{min familjen*, min livet*, gick hennes hemma*}) compared to more advanced levels. 

Despite the corpus being new, it has already been used for several research projects \cite{jantti2011,olars2014,rosare2012}, where  focus varied between linguistic studies, didactic and pedagogic investigations. Rösare \shortcite{rosare2012} investigates what help test takers can get from grammar checkers. Jäntti \shortcite{jantti2011} studies the development of definiteness and adjective agreement in learners of two language groups. Olars \shortcite{olars2014} presents a case study on pragmatics and the problem of communicative writing in a test situation. Several other SweLL-based projects are ongoing. 

\section{Acknowledgements}
\textit{Center for Language Technology} at the University of Gothenburg has financed the work on SpIn, and all the work on converting SweLL digitized essays into browsable texts enriched with linguistic annotation. For the digitization of Tisus, funding was provided by \textit{Ebba Danelius stiftelse för sociala och kulturella ändamål and Granholms stiftelse}, at Stockholm University. 

\section{References}

\bibliographystyle{lrec2016}
\bibliography{lrec2016-ref}

\end{document}